%% file: bare_conf.tex
\documentclass[10pt, conference]{IEEEtran}
\IEEEoverridecommandlockouts
\usepackage{epsfig}
\usepackage{graphicx}
\usepackage{amsmath}
\usepackage{amssymb}
\usepackage{float}
\usepackage{tabularx}
\usepackage{bbm}
\usepackage{breqn}
\usepackage{booktabs}
\usepackage{placeins}

\usepackage[pagebackref=true,breaklinks=true,colorlinks,bookmarks=false]{hyperref}

\ifCLASSINFOpdf
\else
\fi

\begin{document}
%
\title{YOLO-LITE: A Real-Time Object Detection Algorithm Optimized for Non-GPU Computers}


\author{\IEEEauthorblockN{Rachel Huang\footnotemark{*}}
\IEEEauthorblockA{School of Electrical and Computer Engineering\\
Georgia Institute of Technology\\
Atlanta, United States\\
rachuang22@gmail.com}
\thanks{*equal authorship}
\and
\IEEEauthorblockN{Jonathan Pedoeem\footnotemark{*}}
\IEEEauthorblockA{Electrical Engineering\\
The Cooper Union\\
New York, United States\\
pedoeem@cooper.edu}
\and
\IEEEauthorblockN{Cuixian Chen}
\IEEEauthorblockA{Mathematics and Statistics\\
UNC Wilmington\\
North Carolina, United States\\
chenc@uncw.edu}
}


%



\maketitle

\input{abstract.tex}


%
\IEEEpeerreviewmaketitle

\input{introduction.tex}

\input{lit.tex}
\input{experiment.tex}

\input{results.tex}
\input{conclusion.tex}
\input{future.tex}

{\small
\bibliographystyle{IEEEtran}
\bibliography{IEEEexample.bib}

\end{document}

%% file: abstract.tex
\begin{abstract}
 This paper focuses on YOLO-LITE, a real-time object detection model developed to run on portable devices such as a laptop or cellphone lacking a Graphics Processing Unit (GPU). The model was first trained on the PASCAL VOC dataset then on the COCO dataset, achieving a mAP of 33.81\% and 12.26\% respectively. YOLO-LITE runs at about 21 FPS on a non-GPU computer and 10 FPS after implemented onto a website with only 7 layers and 482 million FLOPS. This speed is $\textbf{3.8}\times$ faster than the fastest state of art model, SSD MobilenetvI. Based on the original object detection algorithm YOLOV2, 
YOLO-LITE was designed to create a smaller, faster, and more efficient model increasing the accessibility of real-time object detection to a variety of devices.
\end{abstract}

%% file: introduction.tex
\section{Introduction}
In recent years, object detection has become a significant field of computer vision. The goal of object detection is to detect and classify objects leading to many specialized fields and applications such as face detection and face recognition.
Vision is not only the ability to see a picture in ones head but also the ability to understand and infer from the image that is seen.
The ability to replicate vision in computers is necessary to progress day to day technology. Object detection 
addresses this issue by predicting the location of objects through bounding boxes while simultaneously classifying each object in a given image \cite{schmidhuber2015deep, girshick2014rich, szegedy2013deep}. 

 In addition, with recent developments in technology such as autonomous vehicles, precision and accuracy are no longer the only relevant factors. A model's ability to perform object detection in real-time is necessary in order to accommodate for a vehicle's real-time environment.
An efficient and fast object detection algorithm is key to the success of autonomous vehicles \cite{fridman2017autonomous}, augmented reality devices \cite{akgul2016applying}, and other intelligent systems. 
A lightweight algorithm can be applied to many everyday devices, such as an Internet connected doorbell or thermostat.
Currently, the state-of-the-art object detection algorithms used in cars rely heavily on sensor output from expensive radars and depth sensors. Other techniques that are solely computer based require immense amount of GPU power and even then are not always real-time, making them impractical for everyday applications. 
The general trend in computer vision is to make larger and deeper networks to achieve higher accuracy \cite{szegedy2016rethinking, szegedy2017inception, he2016deep, simonyan2014very}. However, such improvement in accuracy with heavy computational cost  may not be helpful to face the challenge in many real world applications which require real-time performance carried out in a computationally limited platform.   

\begin{figure}[t]
\includegraphics[width=0.99\linewidth]{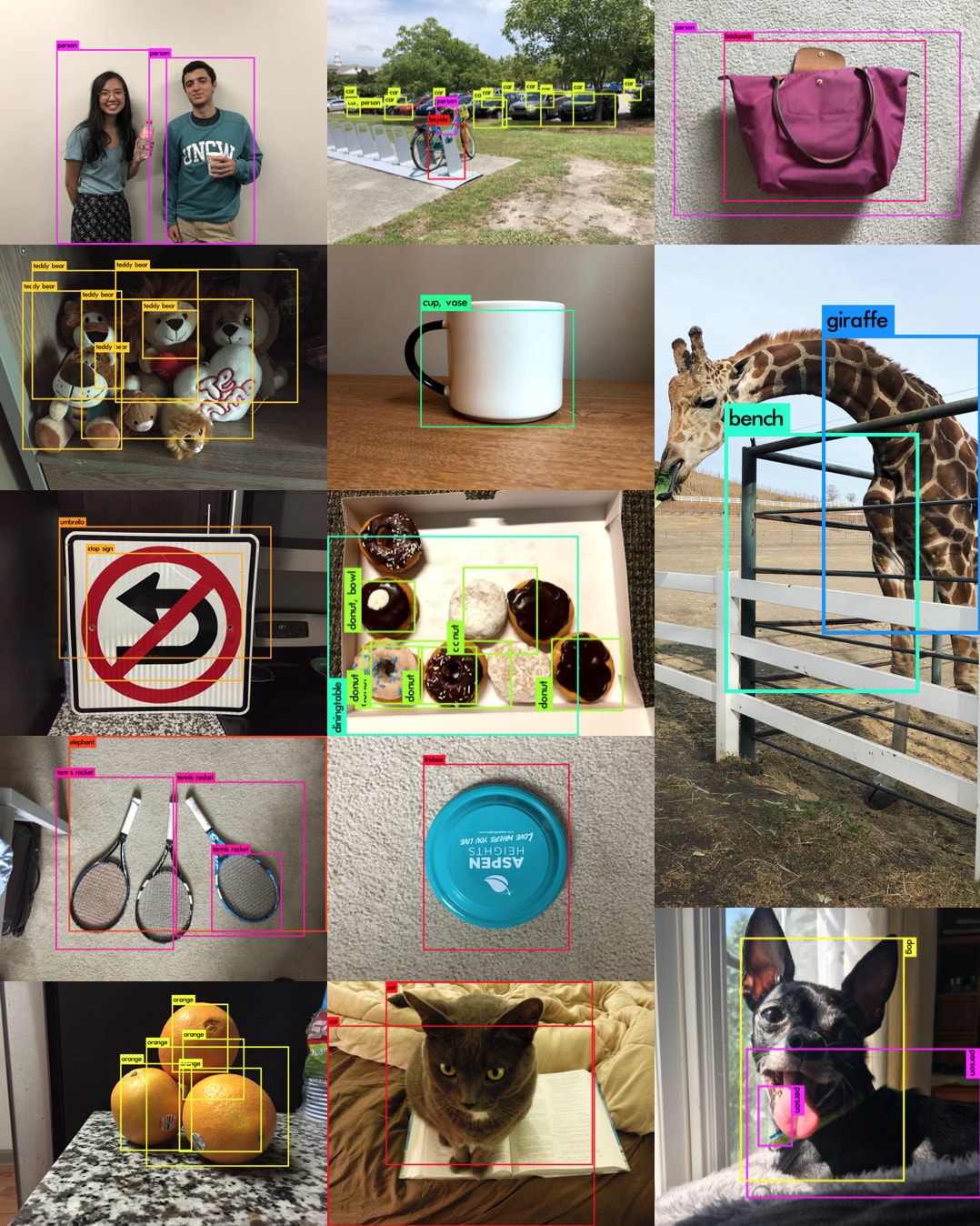}
\caption{Example images passed through our YOLO-LITE COCO model.}
\end{figure}

Previous methods, such as You-Only-Look-Once (YOLO) \cite{redmon2016you} and Regional-based Convolutional Neural Networks (R-CNN) \cite{DBLP:journals/corr/GirshickDDM13}, have successfully achieved an efficient and accurate model with high mean average precision (mAP); however, their frames per second (FPS) on non-GPU computers render them useless for real-time use.  
In this paper, YOLO-LITE is presented to address this problem. Using the You Only Look Once (YOLO) \cite{redmon2016you} algorithm as a starting point, YOLO-LITE is an attempt to get a real time object detection algorithm on a standard non-GPU computer.

%% file: lit.tex
\section{Related Work}
There has been much work in developing object detection algorithms using a standard camera with no additional sensors. State-of-the-art object detection algorithms use deep neural networks.

Convolutional Neural Networks (CNNs) is the main architecture that is used for computer vision. Instead of having fully-connected layers, a CNN has a convolution layer where a filter is convolved with different parts of the input to create the output. The use of a convolution layer allows for relational patterns to be drawn from an input. In addition, a convolution layer tends to have less weights that need to be learned than a fully connected layer as filters do not need an assigned weight from every input to every output.

\subsection{R-CNN}
Regional-based convolutional neural networks (R-CNN) \cite{DBLP:journals/corr/GirshickDDM13} consider region proposals for object detection in images. From each region proposal, a feature vector is extracted and fed into a convolutional neural network. For each class, the feature vectors are evaluated with Support Vector Machines (SVM). Although R-CNN results in high accuracy, the model is not able to achieve real-time speed even with Fast R-CNN \cite{DBLP:journals/corr/Girshick15} and Faster R-CNN \cite{DBLP:journals/corr/RenHG015} due to the expensive training process and the inefficiency of region proposition.

\subsection{YOLO}
You Only Look Once (YOLO) \cite{redmon2016you} was developed to create a one step process involving detection and classification. Bounding box and class predictions are made after one evaluation of the input image. The fastest architecture of YOLO is able to achieve 45 FPS and a smaller version, Tiny-YOLO, achieves up to 244 FPS (Tiny YOLOv2) on a computer with a GPU.

\begin{figure}[H]
\begin{center}
   \includegraphics[width=.65\linewidth]{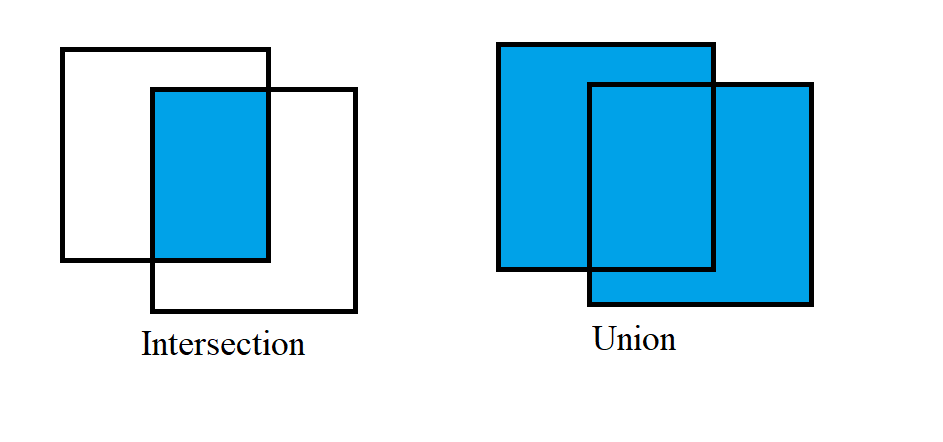}
\end{center}
   \caption{\small Illustration depicting the definitions of intersection and union.}
\label{IOU}
\end{figure}

The idea of YOLO differs from other traditional systems in that bounding box predictions and class predictions are done simultaneously. The input image is first divided into a $\textit{S} 
\times \textit{S}$ grid. Next, \textit{B} bounding boxes are defined in every grid cell, each with a confidence score. Confidence here refers to the probability an object exists in each bounding box and is defined as: 
\begin{align}
C = Pr\left(Object\right)*IOU_{pred}^{truth}
\end{align}
where IOU, intersection over union, represents a fraction between 0 and 1. Intersection is the overlapping area between the predicted bounding box and ground truth, and union is the total area between both predicted and ground truth as illustrated in Figure \ref{IOU}. Ideally, the IOU should be close to 1, indicating that the predicted bounding box is close to the ground truth. 

Simultaneously, while the bounding boxes are made, each grid cell also predicts \textit{C} conditional class probability. The class-specific probability for each grid cell \cite{redmon2016you} is defined as:

\begin{align}
&Pr\left(Class_i|Object\right)*Pr\left(Object\right)*IOU_{pred}^{truth} \notag\\   = &Pr\left(Class_i\right)*IOU_{pred}^{truth}.
\label{yolo_equation}
\end{align}

YOLO uses the following equation below to calculate loss and ultimately optimize confidence:

\begin{align}
Loss = \notag\\
&\lambda_{coord}\sum_{i=0}^{s^2}\sum_{j=0}^{A}\mathbbm{1}_{ij}^{obj}[(b_{x_i}-b_{\hat{x}_i})^2+(b_{y_i}-b_{\hat{y}_i})^2] \notag\\
&+ \lambda_{coord}\sum_{i=0}^{s^2}\sum_{j=0}^{A}\mathbbm{1}_{ij}^{obj}[(\sqrt{b_{{w_i}}}-\sqrt{b_{\hat{w}_i}})^2+(\sqrt{b_{h_i}}-\sqrt{b_{\hat{h}_i}})^2] \notag\\
&+ \sum_{i=0}^{s^2}\sum_{j=0}^{A}\mathbbm{1}_{ij}^{obj}(C_i-\hat{C}_i)^2 \notag\\
&+ \lambda_{noobj}\sum_{i=0}^{s^2}\sum_{j=0}^{A}\mathbbm{1}_{ij}^{noobj}(C_i-\hat{C}_i)^2 \notag\\
&+\sum_{i=0}^{s^2}\mathbbm{1}_{i}^{obj}\sum_{c\in classes}(p_i(c)-\hat{p}_i(c))^2.
\label{loss}
\end{align}

The loss function is used to correct the center and the bounding box of each prediction. Each image is divided into an $S \times S$ grid, with $A$ bounding boxes for each grid. 
The $b_{\textit{x}}$ and $b_{\textit{y}}$ variables refer to the center of each prediction, while $b_{\textit{w}}$ and $b_{\textit{h}}$ refer to the bounding box dimensions. The $\lambda_{coord}$ and $\lambda_{noobj}$ variables are used to increase emphasis on boxes with objects, and lower the emphasis on boxes with no objects. \textit{C} refers to the confidence, and \textit{p(c)} refers to the classification prediction. The $\mathbbm{1}_{ij}^{obj}$ is 1 if the $j^{th}$ bounding box in the $i^{th}$ cell is responsible for the prediction of the object, and 0 otherwise. $\mathbbm{1}_{i}^{obj}$ is 1 if the object is in cell $i$ and 0 otherwise. The loss indicates the performance of the model, with a lower loss indicating higher performance.

While loss is used to gauge performance of a model, the accuracy of predictions made by models in object detection are calculated through the average precision equation shown below: 
\begin{align}
\textit{avgPrecision} = \sum_{\textit{k}=1}^{\textit{n}}P(\textit{k})\Delta\textit{r(k)}.
\end{align}
P(\textit{k}) here refers to the precision at threshold \textit{k} while $\Delta\textit{r(k)}$ refers to the change in recall.

\begin{table}
\footnotesize
\begin{center}
\begin{tabular}{c c c c c c}
\hline
\textbf{Model} & \textbf{Layers} & \textbf{FLOPS (B)} & \textbf{FPS} & \textbf{mAP} & \textbf{Dataset}\\ \hline
YOLOv1  &  26 & not reported  &  45  & 63.4 & VOC\\ 
YOLOv1-Tiny  & 9 & not reported  &  155  & 52.7 & VOC
\\ 
YOLOv2  &  32 & 62.94	&  40 	   & 48.1   & COCO      \\ 
YOLOv2-Tiny& 16  &5.41  &  244     &   23.7   & COCO   \\
YOLOv3& 106  &140.69  &  20     &  57.9   & COCO  \\
YOLOv3-Tiny& 24  &5.56  &  220     &   33.1  & COCO   \\\hline
\end{tabular}
\end{center}
\caption{\small Performance of each version of YOLO.}
\label{YOLOtable}
\end{table}

 The neural network architecture of YOLO contains 24 convolutional layers and 2 fully connected layers. YOLO is later improved with different versions such as YOLOv2 or YOLOv3 in order to minimize localization errors and increase mAP. As seen in Table \ref{YOLOtable}, a condensed version of YOLOv2, Tiny-YOLOv2 \cite{redmon2017yolo9000}, has a mAP of 23.7\% and the lowest floating point operations per second (FLOPS) of 5.41 billion.

When Tiny-YOLOv2 runs on a non-GPU laptop (Dell XPS 13), the model speed decreases from 244 FPS to about 2.4 FPS. With this constriction, real-time object detection is not easily accessible on many devices without a GPU, such as most cellphones or laptops.

\section{YOLO-LITE Architecture}

Our goal with YOLO-LITE is to develop an architecture that can run at a minimum of $\sim 10\;$ frames per second (FPS) on a non-GPU powered computer with a mAP of 30\% on PASCAL VOC. This goal is determined from looking at the state-of-the-art and creating a reasonable benchmark to reach. 
YOLO-LITE offers two main contributions to the field of object detection:
\begin{enumerate}
\item Demonstrates the capability of shallow networks with fast non-GPU object detection applications.
\item Suggests that batch normalization is not necessary for shallow networks and, in fact, slows down the overall speed of the network. 
\end{enumerate}

%% file: experiment.tex

While some works \cite{zhang2017shufflenet,iandola2016squeezenet,howard2017mobilenets} focused on creating an original convolution layer or pruning methods in order to shrink the size of the network, YOLO-LITE focuses on taking what already existed and pushing it to its limits of accuracy and speed. Additionally, YOLO-LITE focuses on speed and not overall physical size of the network and weights. 

Experimentation was done with an agile mindset. Using Tiny-YOLOv2 as a starting point, different layers were removed and added and then trained on Pascal VOC 2007 \& 2012 for about 10-12 hours. All of the iterations used the same last layer as Tiny-YOLOv2. This layer is responsible for splitting the feature map into the $S x S$ grid for predicting the bounding boxes. The trials were then tested on the validation set of Pascal 2007 to calculate mAP. Pascal VOC was used for the development of the architecture, since its small size allows for quicker training. The best performing model was used as a platform for the next round of iterations.

While there was a focus on trying to intuit what would improve mAP and FPS, it was hard to find good indicators. From the beginning, it was assumed that the FLOPS count would correlate with FPS; this proved to be true. However, adding more filters, making filters bigger, and adding more layers did not easily translate  to an improved mAP.

\subsection{Setup}
	Darknet, the framework created to develop YOLO was used to train and test the models. The training was done on a Alienware Aura R7, with a Intel i7 CPU, and a  Nvidia 1070 GPU. Testing for the frames per second were done on a Dell XPS 13 laptop, using Darkflow's live demo example script.
   
\subsection{PASCAL VOC and COCO Datasets}
YOLO-LITE was trained on two datasets. The model was first trained using a combination of PASCAL VOC 2007 and 2012 \cite{PASCAL}. It contains 20 classes with approximately 5,000 training images in the dataset. 

\begin{figure}
\begin{center}
   \includegraphics[width=.80\linewidth]{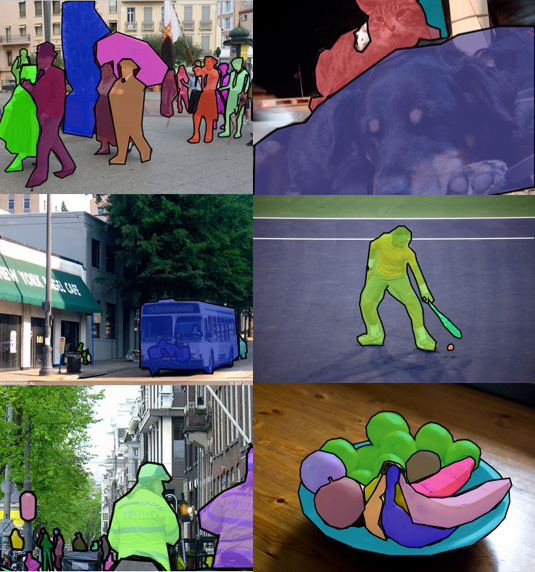}
\end{center}
   \caption{\small Example images with image segmentation from the COCO dataset \cite{COCO}.}
\label{cocodatset}

\end{figure}

The highest performing model trained on PASCAL VOC was then retrained on the second dataset, COCO 2014 \cite{COCO}, containing 80 classes with approximately 40,000 training images. Figure \ref{cocodatset} shows some example images with object segmentation taken from the COCO dataset.

\begin{table}[htbp]
\caption{\small PASCAL VOC and COCO datsets}
\footnotesize
\begin{center}
\begin{tabular}{c c c}
\hline
\textbf{Dataset} & \textbf{Training Images} & \textbf{Number of Classes} \\ \hline
PASCAL VOC 2007 + 2012  & 5,011  & 20 \\ 
COCO 2014  & 40,775  & 80 
\\  \hline
\end{tabular}
\end{center}
\label{datasets}
\end{table}

\begin{table*}[t]
\caption{Results for Each Trial Run on PASCAL VOC.}
\begin{center}
 \begin{tabular}
 {p{0.15\textwidth}p{0.11\textwidth}p{0.15\textwidth}p{0.13\textwidth}p{0.12\textwidth}p{0.11\textwidth}p{0.11\textwidth}}
      \toprule
          Model 			& Layers	& mAP 		& FPS 			& FLOPS	& Time & Loss					\\ \midrule
         Tiny-YOLOv2 (TY)  	&  9 		& 40.48\%  	& 2.4  			&6.97 B        & 12 hours & 1.26             \\  
         TY- No Batch Normalization (NB)  	&  9 		& 35.83\%  	& 3			&6.97 B        & 12 hours &      0.85     \\ 
         Trial 1  			&  7 		& 12.64\%  	& 1.56  		& 28.69 B      & 10 hours & 1.91             \\ 
         Trial 2  			&  9 		& 30.24\%   & 6.94  		& 1.55 B	   & 5 hours &1.37              \\ 
          Trial 2 (NB) 			&  9 		& 23.49\%   & 12.5  		& 1.55 B	   & 6 hours &1.56             \\ 
         Trial 3			&  7  		& 34.59\%   & 9.5  & 482 M    & 10 hours & 1.68             \\ 
       \textbf{Trial 3 (NB)}			&  \textbf{7} 		& \textbf{33.57\%}  	& \textbf{21}  			& \textbf{482 M}    & \textbf{7 hours}   & \textbf{1.64}                   \\ 
       Trial 4		    &  8 		& 2.35\%	& 5.2   		& 1.03 B   	& 10 hours &1.93              \\ 
         Trial 5			&  7 		& .55\%  	& 3.5   		& 426 M    & 7 hours &2.4                      \\ 
	    
         Trial 6			&  7 		& 29.33\%  	& 9.7   		& 618 M    & 11 hours  & 1.91                    \\ 
         Trial 7			&  8 		& 16.84\%  	&  5.7  	    & 482 M    & 7 hours   & 2.3                   \\
         Trial 8			&  8		& 24.22\%  	& 7.8   					& 490 M    & 13 hours & 1.3                      \\ 
         Trial 9			&  7	    & 28.64\%  	& 21   			&  846 M    & 12 hours & 1.5                      \\ 
         Trial 10			&  7	    & 23.44\%  	& 8.2  			& 1.661 B    & 10 hours &1.55                      \\  
         Trial 11			&  7	    & 15.91\%  	& 21   			& 118 M    & 12 hours   &1.35                   \\ 
      	 Trial 12			&  8	    & 26.90\%  	& 6.9   			& 71 M    & 9 hours   &1.74                 \\ 
      	 Trial 12 (NB)			&  8	    & 25.16\%  	& 15.6   			& 71 M    & 12 hours   &1.35                   \\ 
        Trial 13			&  8	    & 39.04\%  	& 5.8  			& 1.083 B   & 11 hours   &1.42                  \\ 
        Trial 13 (NB)			&  8	    & 33.03\%  	& 10.5   			& 1.083 B    & 16 hours   &0.77                   \\ 
      \bottomrule
    \end{tabular}
    \end{center}
    \label{bigtable}
\end{table*}

\begin{table*}[t]
\caption{Architecture of Each Trial on Pascal VOC}
	\begin{center}
    \begin{tabular}{p{0.15\textwidth}p{0.85\textwidth}}
    \toprule
    Trial & Architecture Description \\ \midrule
	TY-NB   & Same architecture at TY, but with no batch normalization. \tabularnewline
    Trial 1 & First 3 layers same as TY. Layer 4 has 512 1 3x3 filters. Layer 5 has 1024 3x3 layers. Layer 6 \& 7 same as the last 2 layers of TY  \tabularnewline
    Trial 2 & Same Architecture as TYV, but input image size shrunk to 208x208  \tabularnewline 
     Trial 2 (NB) & Same architecture as trial 2, but no batch normalization \tabularnewline 
    Trial 3 & First 4 layers same as TYV. Layer 5 has 128 3x3 filters. Layer 6 has 128 3x3 filters. Layer 7 has 256 3x3 filters. Layer 8 has 125 1x1 filters. \tabularnewline 
     \textbf{Trial 3 (NB)} & \textbf{Same architecture as Trial 3, but no batch normalization}   \tabularnewline 
    Trial 4 & Layer 1 5 3x3 filters. Layer 2 5 3x3 filters. Layers 3 16 3x3 filters. Layer 4 64 2x2 filters. Layer 5 256 2x2 filters. Layer 6 128 2x2 filters. Layer 7 512 1x1 filters. Layer 8 125 1x1  filters.  \tabularnewline 
    Trial 5 &  L1 8 3x3 filters. L2 16 3x3 filters. L3 32 1x1 filters. L4 64 1x1 filters. L5 64 1x1 filters. L6 125 1x1 filters.\tabularnewline 
    Trial 6 & Trial 7 is the same as trial 3, but the activation functions were changed to ReLU instead of Leaky ReLU\tabularnewline
    Trial 7 & L1 32 3x3 filters. L2 34 3x3 filters. L3 64 1x1. L4 128 3x3 filters. L5 256 3x3 filters. L6 1024 1x1 filters. L7 125 1x1 filters.  \tabularnewline 
    Trial 8 & Trial 8 is the same as trial 3, but one more Layer before L7 with 256 3x3 filters.   \tabularnewline 
   Trial 9 & Trial 9 is the same as trial 3 (NB), but with the input raised to 300x300.   \tabularnewline 
   Trial 10 & Trial 10 is the same as trial 3 (NB), but with the input raised to 416x416.   \tabularnewline 
   Trial 11 & Trial 11 is the same as trial 3 (NB), but wit the input lowered to 112x112.   \tabularnewline 
  	Trial 12 &  \tabularnewline
    Trial 12 (NB) & Same architecture as trial 12, but no batch normalization \tabularnewline
    Trial 13 & Trial 13 has the same architecture as TY but has one last layer. It does not have layer 8 of TY. \tabularnewline
    Trial 13 (NB) & Same architecture as trial 13,  but no batch normalization \tabularnewline
  
   \bottomrule
	\end{tabular}
	\end{center}
    \label{pascalTrial}
\end{table*}

\raggedbottom
\subsection{Indicators for Speed and Precision}
Table \ref{bigtable} reveals what was successful and what was not when developing YOLO-LITE. The loss which is reported in Table \ref{bigtable}, was not a good indicator of mAP. While a high loss indicates a low mAP there is no exact relationship between the two. This is due to the fact that the losses listed in Equation \ref{loss} are not defined exactly by the mAP but rather a combination of different features.  
The training time, when taken in conjugation with the amount of epochs, was a very good indicator of 
FPS as seen from Trials 3, 6 etc. The FLOPS count was also a good indicator, but given that the FLOPS count does not take into consideration the calculations and time necessary for batch normalization, it was not as good as considering at the epoch rate. 

Trials 4, 5, 8, and 10 showed that there was no clear relationship between adding more layers and filters and improving accuracy. 

\subsection{Image Size}
It was determined when comparing Trial 2 to Tiny-YOLOv2 that reducing the input image size by a half can more than double the speed of the network  (6.94 FPS vs 2.4 FPS) but will also effect the mAP (30.24\% vs 40.48\%). Reducing the input image size means that less of the image is passed through the network. This allows the network to be leaner, but also means that some data was lost. We determined that, for our purposes, it was better to take the speed up over the mAP.  
\subsection{Batch Normalization}

Batch normalization \cite{ioffe2015batch} offers many different improvements namely speeding up the training time. Trials have shown batch normalization improves accuracy over the same network without it. YOLOv2 and v3 have also seen improvements in training and mAP by implementing batch normalization \cite{redmon2017yolo9000,redmon2018yolov3}. While there is much empirical evidence showing the benefits of using batch normalization, it has been found to not be necessary while developing YOLO-LITE. To understand why that is the case, it is necessary to understand what batch normalization is.

Batch normalization entails taking the output of one layer and transforming it to have a mean zero variance and a standard deviation of one before inputting to the next layer. The idea behind batch normalization is that during training using mini-batches it is hard for the network to learn the true ground-truth distribution of the data as each mini-batch may have a different mean and variance. This issue is described as a \textit{covariate shift}. The \textit{covariate shift} makes it difficult to properly train the model as certain features may be disproportionately saturated by activation functions. This is what is referred to as the \textit{vanishing gradient problem}. By keeping the inputs in the same scale, batch normalization stabilizes the network which in turn allows the network to train quicker. During test time, estimated values from training are used to batch normalize the test image. 

As YOLO-LITE is a small network, it does not suffer greatly from \textit{covariate shift} and in turn \textit{vanishing gradient problem}. Therefore, the assumptions made by Ioffe et al. that batch normalization is necessary does not apply. In addition, it seems that the batch normalization calculation that needs to take place in between each layer holds up the network and slows down the whole feedforward process. During the feedfoward each input value has to be updated. For example, the initial input layer of $224\times224\times3$ has over 150k calculations that needs to be made. This calculation happening at every layer builds the time necessary for the forward pass. This has led us to do away with batch normalization. 

\subsection{Pruning}
Pruning is the idea of cutting certain weights based on their importance. It has been shown that a simple pruning method of removing anything less than a certain threshold can reduce the amount of parameters in Alexnet by $9\times$ and VGGnet by $13\times$ with little effect on accuracy  \cite{han2015deep}. 

It has also been suggested \cite{han2015deep} that pruning along with quantization of the weights and Huffman coding can greatly shrink the network size and also speed up the network by 3 or 4 times (Alexnet, VGGnet) . 

Pruning YOLO-LITE showed to have no improvement in accuracy or speed. This is not surprising as the results mentioned in the paper \cite{han2015deep} are on networks that have many fully-connected layers. YOLO-LITE consists mainly of convolutional layers. This explains the lack of results. A method such as the one suggested by Li et al. for pruning convolutional neural networks may be more promising \cite{li2016pruning}. They suggest pruning whole filters instead of select weights.

%% file: results.tex
\section{Results}
\begin{figure*}[th]
\center
\includegraphics[width=0.7\linewidth, keepaspectratio]{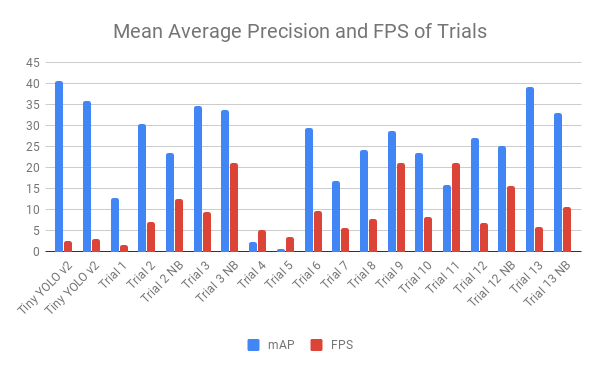}
\begin{center}
\caption{Comparison of trials attempted while developing YOLO-LITE}
\label{compareRes}
\end{center}
\end{figure*}

There were 18 different trials that were attempted during the experimentation phase. Figure \ref{compareRes} shows the mAP and the FPS for each of these trials and Tiny-YOLOv2.

While all the development for YOLO-LITE was done on PASCAL VOC, the best trial was also run on COCO. Table \ref{results} shows the top results achieved on both datasets. 

\begin{table}[H]
	\caption{Results from trial 3 No Batch Normalization (NB)}
	\begin{center}
	\begin{tabular}{p{0.5\linewidth}p{0.25\linewidth}p{0.1\linewidth}} \toprule
  Dataset & mAP& FPS \\ \midrule
  \textbf{PASCAL VOC} & 33.77\% & 21 \\
  \textbf{COCO} & 12.26\% & 21\\ \bottomrule
  \end{tabular}
  \end{center}
  \label{results}
\end{table}

\subsection{Architecture}
Table \ref{YOLO-arch} and Table \ref{trial6-arch} show the architectures for Tiny-YOLOv2 and the best performing trial of YOLO-LITE, Trial 3-no batch. Tiny-YOLOv2 is composed of 9 convolutional layers, a total of 3,181 filters, and 6.97 billion FLOPS.

In contrast, Trial 3-no batch of YOLO-LITE only consists of 7 layers with a total of 749 filters and 482 FLOPS . When comparing the FLOPS of the two models, Tiny-YOLOv2 has $14\times$ more FLOPS than YOLO-LITE Trial 3-no batch. A lighter model containing reduced number of layers enables the faster performance of YOLO-LITE.

\section{Comparison with Other Fast Object Detection Networks}
When it comes to real-time object detection algorithms for non-GPU devices, the competition for YOLO-LITE is pretty slim. YOLO's tiny architecture, which was the starting point for YOLO-LITE has some of the quickest object detection algorithms. While they are much quicker than the bigger YOLO architecture, they are hardly real-time on non-GPU computers ($\sim 2.4$ FPS).

\input{architecture_tables.tex}

Google has an object detection API that has a model zoo with several lightweight architectures \cite{huang2017speed}. The most impressive was SSD Mobilenet V1. This architecture clocks in at 5.8 FPS on a non-GPU laptop with an mAP of 21\%. 

MobileNet \cite{DBLP:journals/corr/HowardZCKWWAA17} uses depthwise separable convolutions, as opposed to YOLO's method, to lighten a model for real-time object detection. The idea of depthwise separable convolutions combines depthwise convolution and pointwise convolution. Depthwise convolution applies one filter on each channel then pointwise convolution applies a 1x1 convolution \cite{DBLP:journals/corr/HowardZCKWWAA17}. This technique aims to lighten a model while maintaining the same amount of information learned in each convolution. The ideas of depthwise convolution in MobileNet potentially explain the higher mAP results from SSD MobileNet COCO. 

 \begin{table}[H]
 \caption{Comparison of State of the Art on COCO dataset}
 	  \small
 	  \centering
      \begin{tabular}{p{0.5\linewidth}p{0.25\linewidth}p{0.1\linewidth}} \toprule  
		Model & mAP& FPS \\ \midrule 
        \textbf{Tiny-YOLOV2} & \textbf{23.7\%} & 2.4 \\
  		\textbf{SSD Mobilenet V1} & 21\% & 5.8 \\	
  		\textbf{YOLO-LITE} & 12.26\% & \textbf{21} \\ \bottomrule
	  \end{tabular}
   	 	\label{sota}
        
 \end{table}
 
	Table \ref{sota} shows how YOLO-LITE compares. YOLO-LITE is \textbf{$ 3.6 \times$} faster than SSD and \textbf{$8.8 \times$} faster than Tiny-YOLOV2.

\subsection{Web Implementation}
After successfully training models for both VOC and COCO, the architectures along with their respective weights files were converted and implemented as a web-based model\footnote{\href{https://reu2018dl.github.io/}{https://reu2018dl.github.io/}} also accessible from a cellphone. Although YOLO-LITE runs at about 21 FPS locally on a Dell XPS 13 laptop, once pushed onto the website, the model runs at around 10 FPS. The FPS may differ depending on the device.

%% file: architecture_tables.tex
\begin{table}{}
\caption{{Tiny-YOLOv2-VOC architecture}}
\begin{center}
    \small
      \begin{tabular}{p{0.25\linewidth}p{0.25\linewidth}p{0.1\linewidth}p{0.1\linewidth} }
      \toprule
          \textbf{Layer} 	& \textbf{Filters}		& \textbf{Size}	& \textbf{Stride} 							\\ \midrule
          Conv1 (C1)  	&  16 		&  $3\times3$  	& 1  			            \\ 
          Max Pool (MP) &   		& $2\times2$ 	& 2  		             \\ 
          C2  			&  32 		& $3\times3$   & 1  		             \\ 
          MP			&    		& $2\times2$   & 2        \\ 
          C3		    &  64 		& $3\times3$	& 1   		        \\
          MP			&   		& $2\times2$  	& 2    		                   \\
	      C4  			&  128 		& $3\times3$  & 1  		             \\ 
          MP			&    		& $2\times2$  & 2            \\ 
          C5		    &  256 		& $3\times3$	& 1   		            \\
          MP			&   		& $2\times2$ 	& 2    		                  \\ 
          C6		    &  512 		& $3\times3$	& 1   		           \\ 
          MP			&   		& $2\times2$  	& 2    		                   \\ 
          C7		    &  1024 	& $3\times3$	& 1   	            \\
          C8		    &  1024 	& $3\times3$	& 1   		          \\ 
          C9		    &  125 		& $1\times1$	& 1   		            \\ 
     	  Region 		 \\
     \hline
    \end{tabular}
  \end{center}
  \label{YOLO-arch}
\bigskip
\caption{YOLO-LITE: Trial 3 Architecture}
    \begin{center}
    \small
      \begin{tabular}{p{0.25\linewidth}p{0.25\linewidth}p{0.1\linewidth}p{0.1\linewidth} }
      \toprule  
         \textbf{Layer} 	& \textbf{Filters}		& \textbf{Size}	& \textbf{Stride}						\\ \midrule
          C1  	&  16 		& $3\times3$  	& 1  		             \\ 
          MP    &   		& $2\times2$	& 2  		            \\ 
          C2  	&  32 		& $3\times3$   & 1  		              \\ 
          MP			&    		& $2\times2$   & 2          \\ 
          C3		    &  64 		& $3\times3$	& 1   		              \\ 
          MP			&   		&$2\times2$  	& 2    	                    \\ 
	      C4  			&  128 		& $3\times3$   & 1  	          \\ 
          MP			&    		& $2\times2$   & 2              \\ 
          C5  			&  128 		& $3\times3$   & 1  	              \\ 
          MP			&    		& $2\times2$   & 2           \\ 
          C6		    &  256 		& $3\times3$	& 1   		            \\ 
          C7		    &  125 		& $1\times1$	& 1   		            \\ 
          Region\\
          \hline
    \end{tabular}
  \end{center}
  \label{trial6-arch}
\end{table}

%% file: conclusion.tex
\section{Conclusion}
YOLO-LITE achieved its goal of bringing object detection to non-GPU computers. In addition, YOLO-LITE offers several contributions to the field of object detection. First, YOLO-LITE shows that shallow networks have immense potential for lightweight real-time object detection networks. Running at 21 FPS on a non-GPU computer is very promising for such a small system. Second, YOLO-LITE shows that the use of batch normalization should be questioned when it comes to smaller shallow networks. Movement in this area of lightweight real-time object detection is the last frontier in making object detection usable in everyday instances. 

%% file: future.tex
\section{Future Work}
Lightweight architectures in general have a significant drop off in accuracy from the original YOLO architecture. YOLOv2 has an mAP of 48.1\% with a decrease to 23.7\% in YOLOv2-Tiny. The YOLO-LITE architecture has a mAP decrease down to 12.16\%. There is a constant tradeoff for speed in lightweight models and accuracy in a larger models. Although YOLO-LITE achieves the fastest mAP compared to state of the art, the accuracy prevents the model from succeeding in real applications such as an autonomous vehicle.
Future work may include techniques to increase the mAP for both COCO and VOC models.

While the FPS for YOLO-LITE is at the necessary level for real-time use with non-GPU computers, the accuracy needs improvement in order for it to be a viable model. 
As mentioned in \cite{redmon,redmon2018yolov3}, pre-training the network to classify on Imagenet has had some good results when transferring the network back to object detection. 

Redmon et al. \cite{redmon2016you} use R-CNN in combination  with YOLO to increase mAP. As previously mentioned, R-CNN finds bounding boxes and classifies each bounding box separately. Once classification is complete, post-processing is used to clean localization errors \cite{redmon2016you}. Although less efficient, an improved version of R-CNN (Fast R-CNN) yields a higher accuracy of 66.9\% while YOLO achieves 63.4\%. When combining R-CNN and YOLO, the model achieves 75\% mAP trained on PASCAL VOC.

Another possible improvement can be to attempt using the feature in YOLOv3 where there are multiple locations for making a prediction. This has helped improve the mAP in YOLOv3 \cite{redmon2018yolov3} and could potentially help improve mAP in YOLO-LITE. 

Filter pruning set out by Li et al. \cite{li2016pruning} can be another possible improvement. While standard pruning did not help YOLO-LITE, pruning out filters can potentially make the network more lean and allow for a more guided training process to learn better weights. While it is not clear if this will greatly improve the mAP, it can potentially speed up the network. It will also make the overall size of the network in memory smaller.

 A final improvement can come from ShuffleNet. ShuffleNet uses group convolution and channel shuffling in order to decrease computation while maintaining the same amount of information during training \cite{zhang2017shufflenet}. Implementing group convolution in YOLO-LITE may improve mAP.

\section{Relevant Links}
More information on the web implementation of YOLO-LITE can be found at \href{https://reu2018dl.github.io/}{https://reu2018dl.github.io/}. For the cfg and weights\ files from training PASCAL VOC and COCO visit \href{https://github.com/reu2018dl/yolo-lite}{https://github.com/reu2018dl/yolo-lite}.

\section{Acknowledgements}
This research was done through the National Science Foundation under DMS Grant Number 1659288. A special thanks to Kha Gia Quach, Chi Nhan Duong, Khoa Luu, Yishi Wang, and Summerlin Thompson for their support.